\documentclass[letterpaper, 10 pt, conference]{ieeeconf}  % Comment this line out if you need a4paper

\usepackage[style=ieee,natbib]{biblatex}
\usepackage{amsmath}
\usepackage{amsfonts}
\usepackage{xspace}
\usepackage{graphicx}
\usepackage{algorithm}
\usepackage{algpseudocode}
\usepackage{caption}

\usepackage{titlesec}
\titlespacing\section{0pt}{3.3pt plus 0pt minus 2pt}{3.3pt plus 0pt minus 2pt}
\titlespacing\subsection{0pt}{1.5pt plus 0pt minus 2pt}{1.5pt plus 0pt minus 2pt}

\IEEEoverridecommandlockouts                              % This command is only needed if 
\newcommand{\ours}[0]{\textsc{SpeedTuning}}

\newcommand{\insertteaser}{
  \centering
  \includegraphics[width=0.99\linewidth]{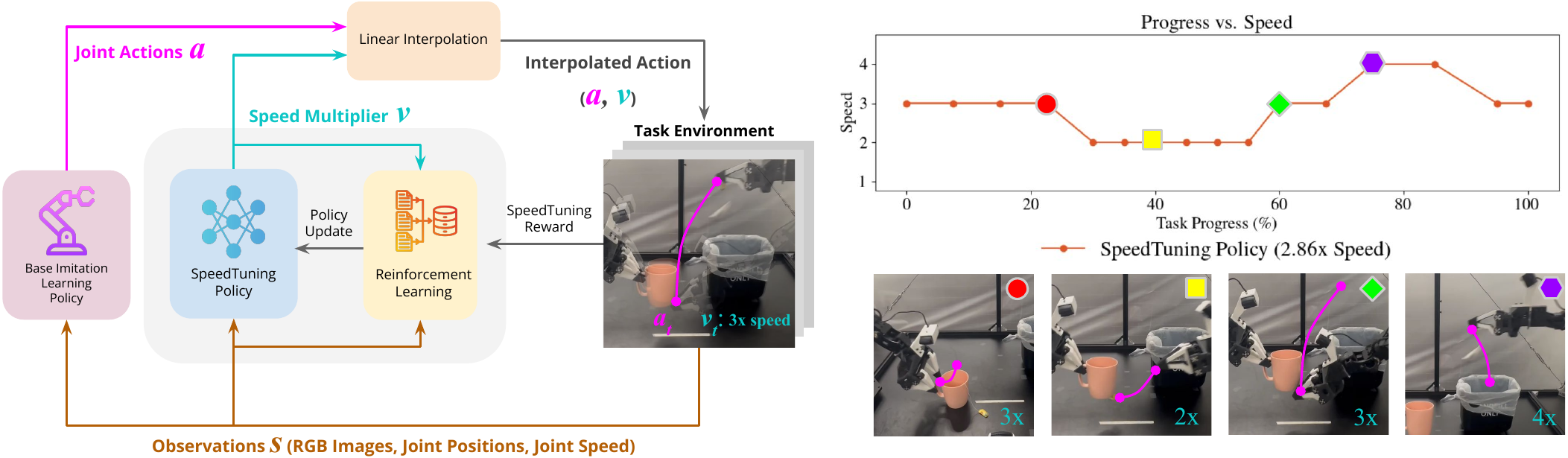} \\
  \captionof{figure}{\ours{} is a reinforcement learning framework designed to optimize both execution speed and task success for learned manipulation policies. The left panel presents a schematic of \ours{} augmenting an imitation learning policy, where it outputs a speed multiplier to modulate the execution of predicted actions. This speed policy is optimized using reinforcement learning. The right panel illustrates performance on the Tea Bag Disposal task, with the upper half depicting speed versus task progress and the lower half showing the speed multipliers predicted at different stages. For critical actions, such as grasping the tea bag (yellow marker), the policy selects a lower speed (2x) to ensure precise timing, whereas for less demanding stages, such as the final phase of discarding the tea bag (purple marker), a higher speed (4x) is applied to expedite execution.}
  \label{fig:teaser}
  \vspace{-1.5pt}
}

\makeatletter
\apptocmd{\@maketitle}{\centering\insertteaser}{}{}%
\makeatother

\title{\LARGE \bf
\ours{}: Speeding Up Policy Execution with Lightweight Reinforcement Learning
}

\author{David D. Yuan, Tony Z. Zhao, Kaylee Burns, and Chelsea Finn\\[0.55em]
{\small Stanford University}
}
\begin{document}

\makeatletter
\def\@pubidpullup{4.5\baselineskip}
\makeatother
\pubid{\parbox{0.96\textwidth}{\centering\scriptsize
\copyright~2025 IEEE. Personal use of this material is permitted. Permission from IEEE must be obtained for all other uses, in any current or future media, including reprinting/republishing this material for advertising or promotional purposes, creating new collective works, for resale or redistribution to servers or lists, or reuse of any copyrighted component of this work in other works.\\
DOI: 10.1109/ICRA55743.2025.11128753}}

\maketitle
\pagestyle{empty}

%%%%%%%%%%%%%%%%%%%%%%%%%%%%%%%%%%%%%%%%%%%%%%%%%%%%%%%%%%%%%%%%%%%%%%%%%%%%%%%%
\begin{abstract}
While learned robotic policies hold promise for advancing generalizable manipulation, their practical deployment is often hindered by suboptimal execution speeds. Imitation learning policies are inherently limited by hardware constraints and the speed of the operator during data collection. In addition, there are no established methods for accelerating policies learned via imitation, and the empirical relationship between execution speed and task success remains underexplored. To address these issues, we introduce \ours{}, a reinforcement learning framework specifically designed to enhance the speed of manipulation policies. \ours{} learns to predict the optimal execution speed for actions, thereby complementing a base policy without necessitating additional data collection. We provide empirical evidence that \ours{} achieves substantial improvements in execution speed, exceeding 2.4x speed-up, while preserving an adequate success rate compared to both the original task policy and straightforward speed-up methods such as linear interpolation at a fixed speed. We evaluate our approach across a diverse set of dynamic and precise tasks, including pouring, throwing, and picking, demonstrating its effectiveness and robustness in enhancing real-world robotic manipulation. Videos and code are available at \url{https://daivdyuan.github.io/speed-tuning/}

\end{abstract}

%%%%%%%%%%%%%%%%%%%%%%%%%%%%%%%%%%%%%%%%%%%%%%%%%%%%%%%%%%%%%%%%%%%%%%%%%%%%%%%%
\section{INTRODUCTION}

% add more problem statement to the first paragraph, then transition into discussion of problems learning speed in IL and the new framework we introduce

% put the problem statement at the beginning of section 4 because we frame it as a contribution

% add uniform interpolation baseline to method after introducing the problem statement. also discuss limitations, which can motivate our main approach

% include algorithm box?

% P1 agile tasks represent the lionshare of real-world manipulation tasks
Speed is critical in real-world robotic tasks, as faster policies can increase throughput and enhance the user experience. 
%For example, in surgical robotics, precise and dynamic manipulation is essential for tasks like suturing, where movements must be both fast and coordinated to avoid tissue damage while maintaining efficiency.  
For example, in robotic assembly, high-speed and precise manipulation is essential for tasks like inserting screws or fitting components, where faster execution directly boosts production efficiency.
Even though speed is essential in many real-world settings, imitation learning methods largely ignore speed both when training and evaluating a policy.
\pubidadjcol

% Dynamic, high-speed tasks represent the lion's share of real-world applications of robotic manipulation. Many consumer applications require dynamic movements, such as placing a tea-bag into a cup, tossing laundry into a basket, or shaking crumbs off of a towel. Execution speed is critical in completing these movements successfully. In factory settings, where floor space is limited and high throughput is required, speed of execution is important in maintaining throughput. Robots must at least match the speed of the other parts of the assembly line to prevent bottlenecks in production. Slower robots leads to longer cycle times, reduced output, and increased production costs. Even though speed is essential in many real-world settings, imitation learning methods largely ignore speed both when training and evaluating a policy.

% P2 a smattering of important history in agility and then more detail about why this is so difficult for imitation learning
Training robots to perform tasks quickly and successfully through imitation is challenging. Imitation learning is typically concerned with matching the behavior in the demonstration dataset, while it is hard to collect fast and accurate teleoperation data even from experts. Take ALOHA~\cite{zhao2023learning}, a widely adopted and performant teleoperation hardware as an example, cutting a piece of tape and sticking it to a box takes more than 15 seconds, and putting a velcro shoe on a foot takes more than 22 seconds. In contrast, humans can perform either of these tasks in under 5 seconds, so simply matching the speed of demonstrations won't result in ``human-level'' performance. Moreover, many real-world tasks are dynamic, requiring precise coordination of actions over time. For instance, tasks like tossing and pouring demand fine velocity control, while actions such as placing a tea bag into a cup require careful timing to synchronize with the bag's swing. Consequently, naive approaches—such as increasing the underlying speed of the robot by a constant multiplier—are insufficient, as they fail to account for the nuanced relationship between speed and task success.
% 

% P3 what are the desiderata of a speed-up system for imitation learning? why is this question sort of interesting?
An ideal speed-up system for imitation learning would (1) not require modifications to demonstration collection mechanisms, (2) retrofit existing policies to work at faster speeds, and (3) learn the interaction between execution speed and policy success. Modeling the relationship between speed and performance is especially important in dynamic tasks, where the speed may need to be increased or decreased depending on the stage of the task. For example, when pulling a tea-bag out of a cup, most of the episode can be accelerated, but grasping the tag of the tea-bag must be timed with the swing of the string, as shown in Fig.~\ref{fig:teaser}.

% P4 concrete description of the speed up approach
We present \ours, a method for accelerating manipulation policies by learning a ``speed-policy'' that predicts a speed factor for a chunk of future actions. The speed-policy is built on top of a base imitation learning policy. As actions are predicted in the environment, the speed policy takes in the observation history and outputs the velocity of the subsequent action. The speed policy is trained with reinforcement learning and optimizes a task and speed reward. To make this practical and compatible with modern imitation learning methods, most of which predict action chunks $a_{t:t+k}$, we linearly interpolate predicted chunks by the factor predicted by the speed policy. Because the speed policy is trained entirely with reinforcement learning, this acceleration process requires no extra data-collection.

% P5 exciting findings
We evaluate \ours{} across a diverse set of dynamic tasks, including pouring, throwing, and picking, in both simulated and real-world environments. We compare our method against reinforcement learning from scratch and a naive speed-up baseline that uniformly accelerates actions. The results demonstrate that \ours{} achieves substantial speed-ups (over 2.4x) compared to the baseline imitation-learning policies. Additionally, we conduct ablation studies to analyze the impact of various design choices, such as the choice of task policy, the use of image observations, and the specific image encoder employed, further validating the effectiveness and robustness of our approach.
% P6 synopsis of findings

\section{Related Work}

\textbf{Imitation learning.} Imitation learning provides a straightforward way to acquire robotic manipulation skills by learning from expert demonstrations~\citep{Pomerleau1988ALVINNAA,Schaal1999IsIL}. Our work builds on a recent paradigm for imitation learning that combines a puppeteering setup~\citep{Hulin2011TheDB,Katz2018ALC,Schwarz2021NimbRoAI,wu2023gello}, transformer architecture~\citep{Vaswani2017AttentionIA}, and action chunking to learn dexterous skills~\citep{zhao2023learning,chi2023diffusion}. Recent advancements in imitation learning highlight the surprising effectiveness of imitation learning for rapid skill acquisition~\citep{Florence2021ImplicitBC,Jang2022BCZZT,chi2023diffusion}. However, the quality of learned skills remains inherently tied to the quality of demonstrations, which can be limited by unintuitive teleoperation interfaces and demonstrator sub-optimality~\citep{kumar2022should}. Furthermore, the current paradigm lacks mechanisms to modify demonstrations or learn policies post-hoc that better optimize for task demands. We focus specifically on optimizing for execution speed and introduce a novel framework to accelerate existing imitation-learned policies for higher-speed manipulation tasks without requiring additional demonstrations or sacrificing performance.

\textbf{Reinforcement learning in robotics.}
A reinforcement learning (RL) agent takes actions in an environment to maximize cumulative reward~\citep{Watkins1992Qlearning,suttonandbarto1998,Sutton1999PolicyGM,Konda1999ActorCriticA}. RL in robotic environments~\citep{Lillicrap2015ContinuousCW,Gu2016DeepRL,Andrychowicz2018LearningDI} evades the issues of sub-optimal demonstration data, but state-of-the art RL algorithms still require extensive real-world interactions to learn complex control policies from scratch~\citep{Ibarz2021HowTT}. Roboticists have successfully fine-tuned policies pre-trained with imitation learning~\citep{Julian2020NeverSL,Yang2023RobotFM}. Unlike these works, we train a new policy that predicts the \textit{velocity} at which the base policy runs.

\textbf{Fast robot manipulation.} 
The pursuit of fast robot manipulation, particularly in dynamic tasks, has traditionally centered around optimizing the robot's hardware or integrating speed into the reward function. These approaches have been successful in highly dynamic scenariossuch as table tennis~\citep{Bchler2020LearningTP,DAmbrosio2024AchievingHL}. Dynamic manipulation tasks, such as cable manipulation~\citep{Yamakawa2009KnottingMO,Yamakawa2010MotionPF,Yamakawa2013DynamicHK,Zhang2020RobotsOT}, throwing objects~\citep{zeng2019tossingbot,Ha2021FlingBotTU}, and high-throughput pick-and-place~\citep{Nabat2005Par4VH,Han2019TowardFA,Pham2018CriticallyFP} also benefit from speed-focused objectives and hardware solutions. Unlike these works, \ours{} is compatible with many base policy learning methods, which means that it can be incorporated on top of already performant robotic policies.

\section{Speed Tuning}
We introduce \ours{}, a method for accelerating manipulation policies learned from demonstrations while optimizing for both execution speed and task success. We achieve this by decoupling the policy into two components: a task policy trained with imitation learning to mimic expert demonstrations, which is summarized in Section~\ref{sec:method-prelim}, and a speed policy trained with reinforcement learning to predict optimal action velocities that balance speed and success. The \ours{} objective is described in Section~\ref{sec:method-objective}. We then present a baseline approach in Section~\ref{sec:method-interpolation} that uniformly accelerates actions to establish the empirical trade-off between speed and success. In Section~\ref{sec:method-ours}, we detail how \ours{} uses RL to predict the velocity of executed actions based on current state to achieve a better combined success and speed than naive speed-up techniques.

\subsection{Preliminaries}
\label{sec:method-prelim}
We assume access to a set of expert demonstrations $\mathcal{D}=\{(\tau_1, \ldots, \tau_
N)\}$ where $\tau_n$ is a trajectory defined as a sequence of state-action pairs: $\tau = [(s_1, a_1) \ldots (s_T, a_T)]$. Under the standard imitation learning paradigm, learning a policy, $\pi$, parameterized by $\theta$ is a matter of minimizing the below loss function to maximize the likelihood of the expert actions:
\begin{align}
    L_\theta &= \mathbb{E}_{\mathcal{D}}\left[d(\pi_\theta(s_i), a_i)\right]
\end{align}

where $d$ is any distance metric, usually Manhattan or Euclidean.

These demonstrations are considered ``expert'' in that they are sampled from a behavior policy optimizing for task reward, $r_{\text{task}}$, which we assume to be a binary reward indicating whether the task was completed during a given transition. This reward signal can be derived either from the structure of the environment or through human-in-the-loop feedback, particularly in real-world reinforcement learning scenarios where automated reward formulation is insufficient.
\begin{align}
    r_{\text{task}}(s_t, a_t) &= {1}_{\{s_{t+1} \in S_\text{success}\}}
\end{align}

Absent from this framework is an understanding of how to speed up policy execution. Typically, we assume actions are executed at a constant velocity, $v$. In the next section, we consider an updated objective where the velocity of actions is a component of the action space. 
% In our setting, we also assume access to a ground-truth task reward $r_{\text{task}}$. Many learning frameworks have been proposed to learn from a combined demonstration set and task reward, but for our purposes 

\begin{algorithm}[t!]
\caption{\ours{} Training}\label{alg:st-train}
\begin{algorithmic}[1] % The "[1]" adds line numbers
\footnotesize

\State \textbf{Given:} 
Demonstration dataset $\mathcal{D_\text{task}}$,  speed reward function $r_{\text{speed}}(v)$ speed reward weight $\alpha$, frame skip constant $k_{\text{skip}}$, frame stack constant $k_{\text{stack}}$, action chunk size $k_c$

\State Initialize task policy $\pi_\theta(a_{t:t+k_c} | s_t)$
\State Initialize speed policy $\pi_\varphi(v_{t} | s_t)$
\State Initialize replay buffer $\mathcal{D}_{\text{replay}}$
\State Initialize observation buffer $\mathcal{O}$

\State \textbf{Train} the task policy $\pi_\theta(a_{t:t+k_c} | s_t)$ using the task dataset $\mathcal{D}_\text{task}$

\For{each episode}
    \State Clear observation buffer $\mathcal{O} = []$
    \For {each timestep $t$}
        \State Obtain current observation $s_{(t, 0)}$
        \State Append current observation to buffer $\mathcal{O} = \mathcal{O} \cup s_{(t, 0)}$
        \State Predict the speed $v_t = \pi_\varphi(\mathcal{O}[-k_\text{stack}:])$
        \State Predict the task action sequence $A_t := a_{t:t+k_c} = \pi_\theta(s_{(t, 0)})$
        \State Interpolate the action based on speed $a^{(A_t, v_t)}$
        \For {iteration $i = 1, 2, \dots, k_\text{skip}$}
            \State Apply interpolated action $a'_i := a_i^{(A_t, v_t)}$ 
            \State Obtain next observation $s_{(t, i)}$ 
            \State Append new observation to buffer $\mathcal{O} = \mathcal{O} \cup s_{(t, i)}$
            \State Obtain task reward $ r_{\text{task}}(s_{(t, i-1)}, a'_i)$
            
            \State Compute the combined reward:
            \[
            r_{ST} = \alpha \cdot r_{\text{speed}}(v_t) + r_{\text{task}}(s_{(t, i-1)}, a'_i)
            \]
            \State Store transition in the replay buffer:
            \[
            \mathcal{D}_{\text{replay}} = \mathcal{D}_{\text{replay}} \cup (s_{(t, i-1)}, a'_i, r_{ST}, s_{(t, i)})
            \]
        \EndFor
    \EndFor
    \State \textbf{Sample} mini-batches from the replay buffer $\mathcal{D}_{\text{replay}}$ to update the speed policy $\pi_\varphi(v_t | s_t)$
\EndFor
\end{algorithmic}
\end{algorithm}

\subsection{\ours{} Objective}
\label{sec:method-objective}
To extend our framework to account for action execution speed, we introduce an updated objective function that incorporates velocity as a component of the action space. We define the objective $J(\psi)$ as the expected cumulative reward over trajectories sampled from a policy $\pi_{\psi}(a_t, v_t | s_t)$, where the reward at each time step $t$ consists of a weighted speed reward $\alpha \cdot r_{\text{speed}}(v_t)$ and the task reward $r_{\text{task}}(s_t, a_t)$. We restrict the selection of velocities to a discrete set: $V = \{v_1, v_2, \ldots, v_K\}$ to simplify policy optimization.
\begin{align}
J(\psi) &= \mathbb{E}_{\tau \sim \pi_{\psi}} \left[\sum_{t=1}^{T} \left( \alpha \cdot r_{\text{speed}}(v_t) + r_{\text{task}}(s_t, a_t) \right) \right] \label{eq:dual_reward}
\end{align}

By introducing the weighting term $\alpha \geq 0$ to the velocity reward, we can adjust the trade-off between execution speed and task success. This allows us to describe the Pareto curve of optimality, optimizing for either faster execution (higher $\alpha$) or higher task success rate (lower $\alpha$). 

\subsection{Factorizing Task and Speed}
We decompose the execution policy into two distinct sub-policies: a task policy, \(\pi_\theta(a_t|s_t)\), and a speed policy, \(\pi_\varphi(v_t|s_t)\). This not only makes optimizing for speed and success more tractable, but also enables us to accelerate any parameterization of $\pi_\theta$. In our method, the task policy is trained to regress actions with imitation learning. For the speed policy, we explore both constant and non-linear speed policies, which are described in the next two sections.

\subsection{Constant Velocity}
\label{sec:method-constant}
In scenarios where the velocity $v$ is constant, the objective simplifies to:
\begin{align}
J(\psi) &= \alpha \cdot v T + \mathbb{E}_{\tau \sim \pi_{\psi}} \left[\sum_{t=1}^{T} r_{\text{task}}(s_{t}, a_t) \right].
\end{align}
This formulation enables us to explore different points along the Pareto frontier by varying the weighting term $\alpha$, thereby balancing the trade-off between speed and task performance. However, in practice, more complex non-linear speed reward functions calculated at each transition are often preferred, which we discuss in the next section.

\begin{figure*}[t!]
    \centering
    \includegraphics[width=0.97\linewidth]{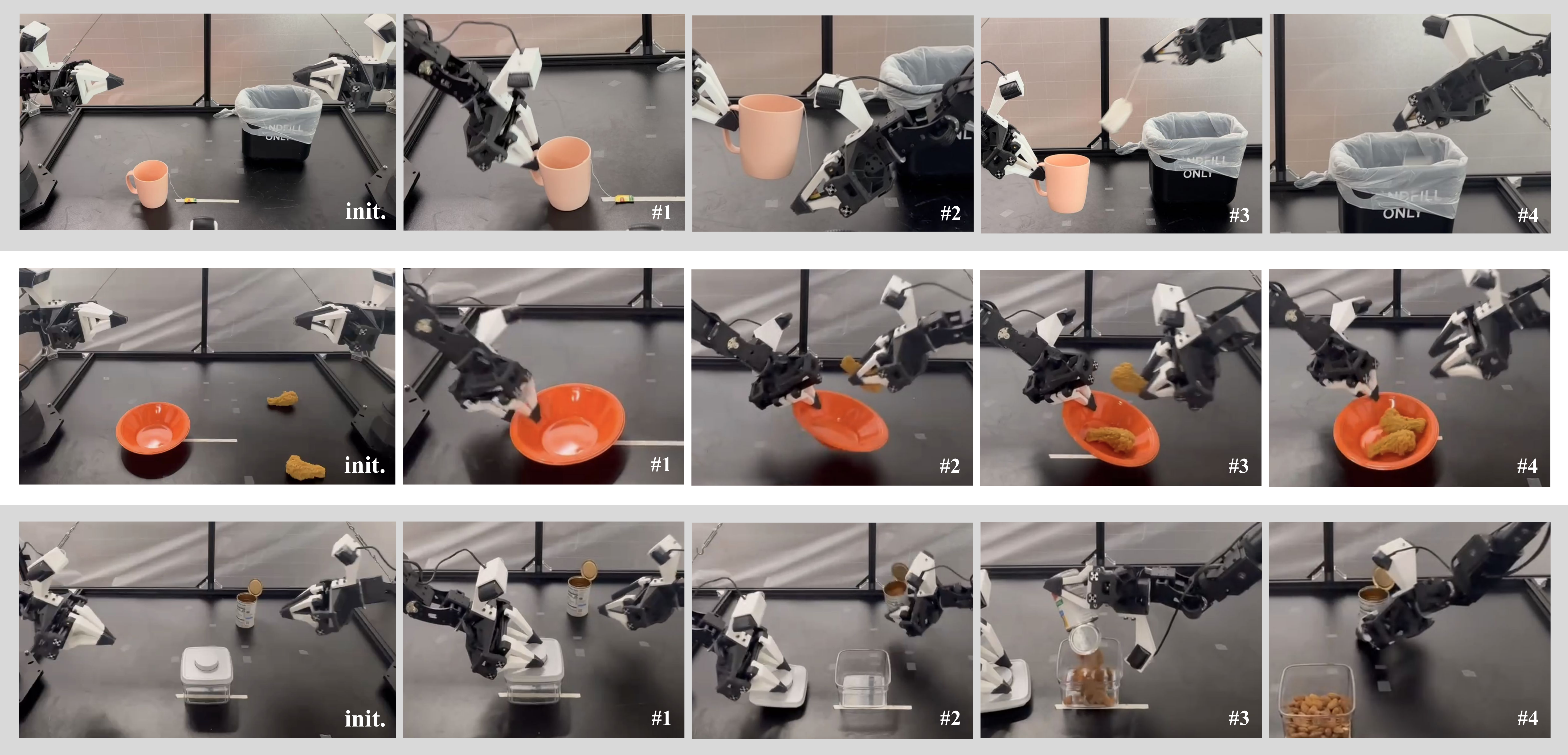}
    \caption{\small Initialization and key actions for three real-world tasks (top to bottom): Tea Bag Disposal, Food Preparation, and Almond Pouring.}
    \label{fig:task_reak}
    \vspace{-0.7cm}
\end{figure*}

\subsection{\ours{}: Learning a Speed-Adaptive Policy}
\label{sec:method-ours}
While uniformly increasing the velocity provides a simple method for accelerating action sequences, it lacks adaptability to different task contexts. Our aim is to develop a more general closed-loop policy where acceleration is conditioned on the current state. For instance, in a pick-and-place scenario, an optimal speed policy would likely slow down during critical actions, such as grasping or placing, to ensure successful execution, and speed up during less critical movements to maximize overall efficiency.

To achieve this, \ours{} seeks to balance swift task execution with maintaining a high success rate. Specifically, we use reinforcement learning with the objective in Equation~\ref{eq:dual_reward} to train a speed policy \(\pi_\varphi(v_t|s_t)\), which predicts an optimal speed \( v_t \) for a fixed task policy $\pi_\theta$:
\begin{align}
J(\varphi) = \mathbb{E}_{\substack{a_t \sim \pi_{\theta} \\ v_t \sim \pi_\varphi}} \left[\sum_{t=1}^{T} \left( \alpha \cdot r_{\text{speed}}(v_t) + r_{\text{task}}(s_t, a_t) \right) \right] \label{eq:split_objective}
\end{align}
To optimize the objective in Equation~\ref{eq:split_objective} we employ Rainbow DQN~\cite{hessel2018rainbow}, which is a value-based off-policy algorithm for discrete action spaces, because of its strong performance and high sample efficiency. Adapting the Rainbow DQN objective for our \ours{} gives the following optimal speed policy and Bellman equation:
\begin{align}
\pi^*_\varphi(s_t, a_t) &= \arg\max_{v_t} Q^*(s_t, a_t, v_t), \label{eq:optimal_speed_policy} \\
r_{\text{ST}}(s_t, a_t, v_t) &= \alpha \cdot r_{\text{speed}}(v_t) + r_{\text{task}}(s_t, a_t), \label{eq:reward_function} \\
Q^{\pi_\varphi}(s_t, a_t, v_t) &= r_{\text{ST}}(s_t, a_t, v_t) + {} \nonumber \\
&\quad \gamma \, \mathbb{E}_{s_{t+1}} \left[ Q^{\pi_\varphi}\left(s_{t+1}, a_{t+1}, v_{t+1} \right) \right], \label{eq:bellman_equation}
\end{align}

where \( \gamma \) is the discount factor.

We made several careful design choices for the speed network. As described in Section~\ref{sec:method-objective}, we limit the output speed \( v \) to a discrete set of possible values. We also introduce an additional hyperparameter, $\beta$, to further scale the speed reward: \( r_{\text{speed}}(v) = v^{\beta} \). To manage complexity in long-horizon tasks, we use a frame skip constant \( k_{\text{skip}} \in \mathbb{Z^+} \), querying the speed policy only once every \( k_{\text{skip}} \) steps, while continuing executing the accelerated policy with the latest speed between queries. To better preserve the Markovian properties, we adopt a frame stack of \( k_{\text{stack}} \) recent observations as input.
However, the most important design decision we make is to predict an action \textit{chunk} at each time step. This makes the implementation of the velocity change more straightforward and reduces the effective horizon of the task. We describe the concrete interpretation of velocity when predicting action chunks in the next section.

\subsection{Interpolation over Action Chunks}
\label{sec:method-interpolation}
Instead of manually scaling the velocity at which actions are run, we assume that the base policy predicts chunks of actions of size $k$ for joint positions:  \( \pi_{\theta}(A_t|s_t) \) where \( A_t = a_{t:t+k} \). To modulate the execution speed of actions generated by the task policy, we employ \textit{linear temporal interpolation}. 

With linear temporal interpolation, the \( i \)-th accelerated action at speed \( v \) for time step \( t \) is denoted as \( a^{(A_t, v)}_i \), with \( a^{(A_t, v)}_1 \) selected as the next action to be executed at the current time step.

Formally, we define linear temporal interpolation over a function \( f: \mathbb{Z} \mapsto \mathbb{R}^n \) with a step length of \( v \) at point \( t \) as:
\begin{equation}
\small
\text{interp}_{f, v}(t) = f\left(\left\lfloor v t \right\rfloor\right) + \frac{v t - \left\lfloor v t \right\rfloor}{v} \big( f\left(\left\lfloor v t + 1 \right\rfloor\right) - f\left(\left\lfloor v t \right\rfloor\right)\big).
\end{equation}
This linear temporal interpolation effectively accelerates the execution of the function \( f \) by a factor of \( v \). For simplicity, we denote a trajectory of actions as \( A = a_{0:k} \). The temporal action function \( f^{(A)}_a: \mathbb{Z} \mapsto \mathbb{R}^n \) is defined as:
\begin{equation}
f^{(A)}_a(i) = a^{(A)}_i, \quad 0 \leq i < k+1.
\end{equation}
The accelerated action at speed \( v \) over the trajectory \( A \) at step \( i \) is then:

\begin{equation}
a^{(A, v)}_i = \text{interp}_{f^{(A)}_a, v}(i), \quad 0 \leq i \leq \left\lfloor \frac{k+1}{v} \right\rfloor.
\end{equation}

\begin{figure*}
    \centering
    \includegraphics[width=.30\linewidth]{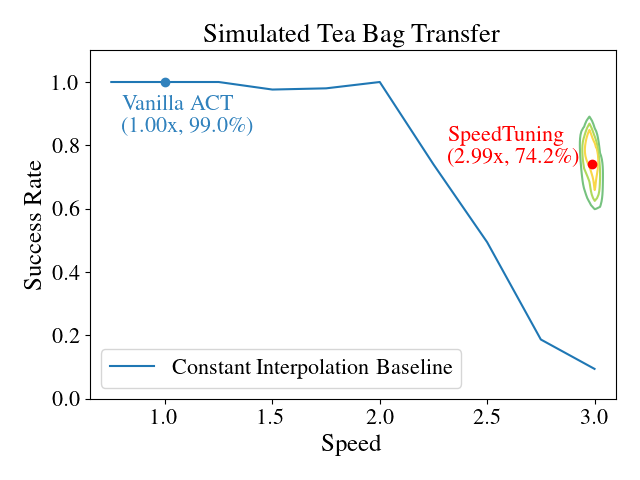}
    \includegraphics[width=.30\linewidth]{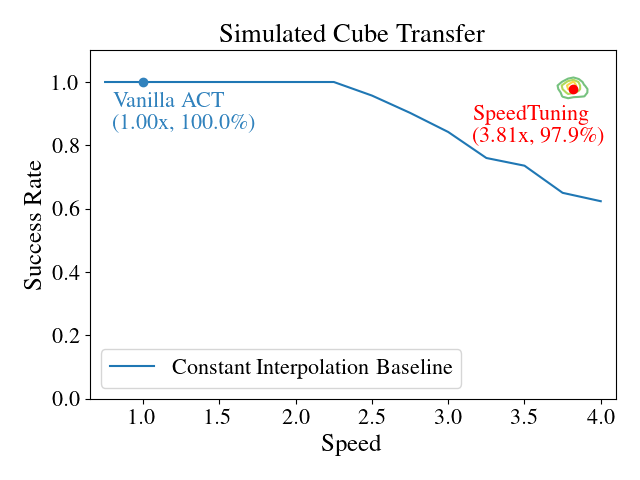}
    \includegraphics[width=.30\linewidth]{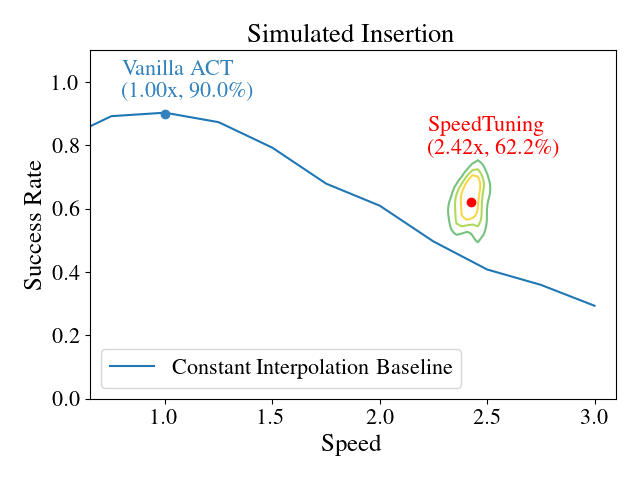}
    \includegraphics[width=.30\linewidth]{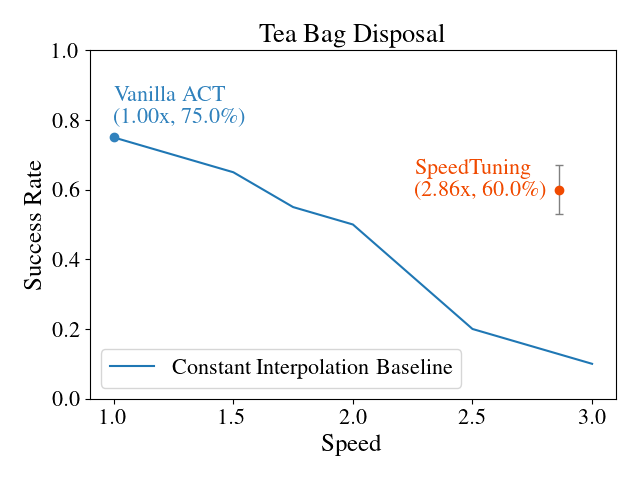}
    \includegraphics[width=.30\linewidth]{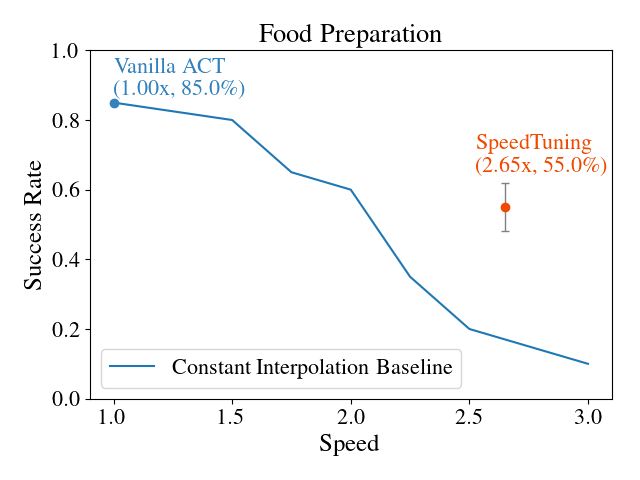}
    \includegraphics[width=.30\linewidth]{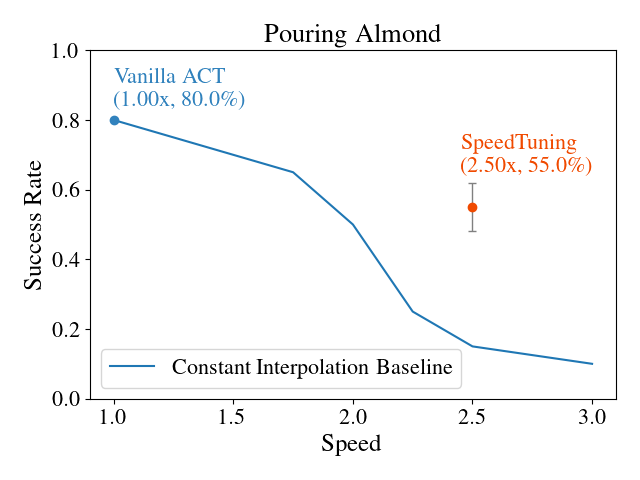}
    \caption{Results showing the \ours{} policy on simulated (top row) and real world (bottom row) tasks. For simulated tasks, the baseline curve is obtained by evaluating 100 episodes across various speeds at 0.1 intervals, while the distribution and mean of the \ours{} policy is calculated over 2000 episodes. For the real-world tasks, the baseline curve is obtained by evaluating 20 episodes across various speeds at 0.25 intervals, while the mean and error of the \ours{} policy is calculated over 20 episodes. Due to the sample efficiency constraints of reinforcement learning in real-world settings, the policy was not trained to full convergence, which may account for the slightly lower observed success rate.}
    \label{fig:res_sim}
    \label{fig:res_real}
    \vspace{-0.5cm}
\end{figure*}

% \begin{figure*}
%     \centering
%     \includegraphics[width=.30\linewidth]{figures/results/tea_bag_sim_result.png}
%     \includegraphics[width=.30\linewidth]{figures/results/cube_sim_result.png}
%     \includegraphics[width=.30\linewidth]{figures/results/insertion_sim_result.png}
%     \caption{\small Results for the simulated tasks, showing the distribution and mean value of the \ours{} policy. The baseline curve is obtained by evaluating 100 episodes across various speeds at 0.1 intervals, while the \ours{} policy is evaluated over 2000 episodes.}
%     \label{fig:res_sim}
% \end{figure*}
% % \vspace{-1em}

% \begin{figure*}
%     \centering
%     \includegraphics[width=.30\linewidth]{figures/results/tb_real_result.png}
%     \includegraphics[width=.30\linewidth]{figures/results/fp_real_result.png}
%     \includegraphics[width=.30\linewidth]{figures/results/pa_real_result.png}
%     \caption{\small Results for the real-world tasks, with the mean value and error margins of the \ours{} policy.  The baseline curve is obtained by evaluating 20 episodes across various speeds at 0.25 intervals, while the \ours{} policy is evaluated over 20 episodes. Due to the sample efficiency constraints of reinforcement learning in real-world settings, the policy was not trained to full convergence, which may account for the slightly lower observed success rate.}
%     \label{fig:res_real}
% \end{figure*}

\begin{figure*}
    \centering
    \includegraphics[width=0.92\linewidth]{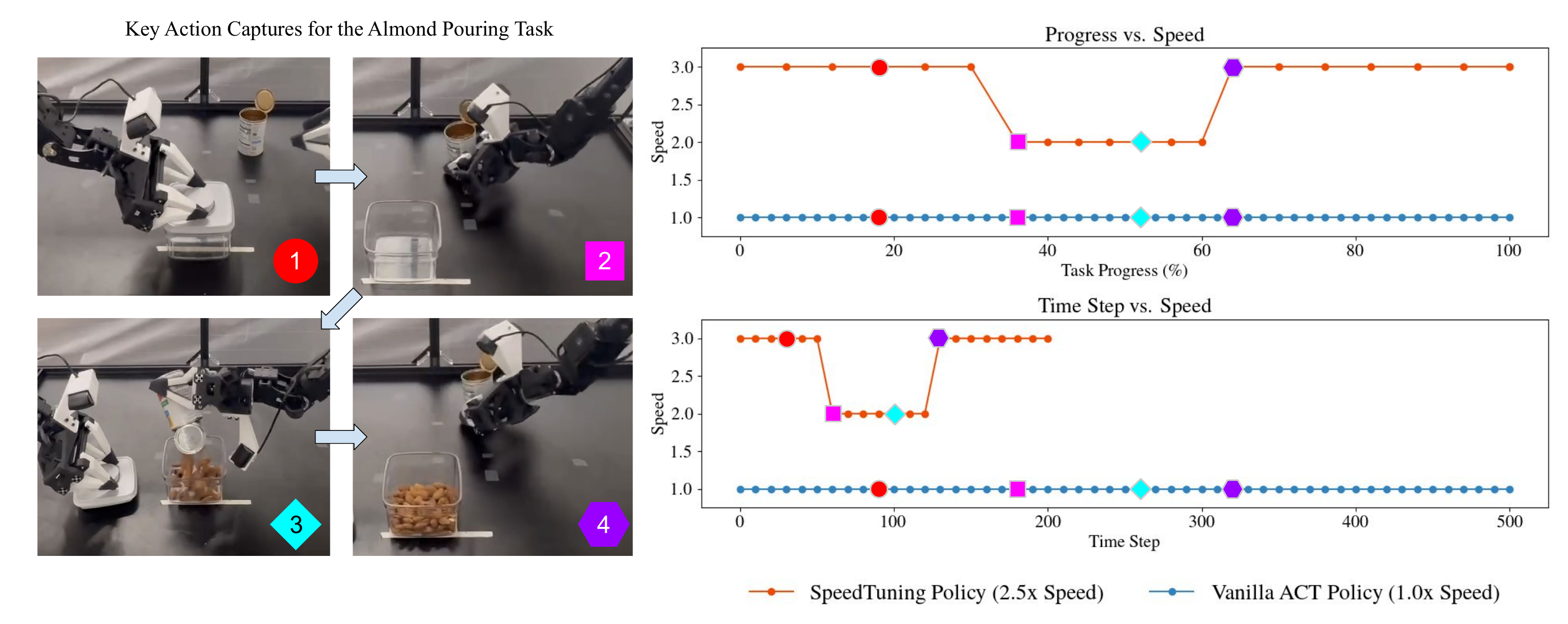}
    \caption{\small Comparison of the \ours{} and vanilla ACT policy trajectories for the Almond Pouring task, featuring key action captures and speed-up values throughout the task. The results highlight \ours{}'s task understanding, significantly accelerating overall execution by 2.5x (200 time steps vs. 500 time steps), while deliberately slowing down during the critical, dynamic pouring phase.}
    \label{fig:traj_real}
\vspace{-0.5cm}
\end{figure*}

In summary, this interpolation method allows us to integrate the speed policy's output, $v$, with the action chunks generated by a task policy by decreasing the effective chunk-size down to $\frac{k+1}{v}$.

\section{Experiments}

In this section, we present a comprehensive evaluation of \ours{} to demonstrate its effectiveness in accelerating policy execution. In the following sections, we discuss the implementation details and describe our simulated and real-world dynamic manipulation tasks. Then, we present our key results, which show that \ours{} achieves significant speed-ups while maintaining solid policy performance. Finally, we ablate the design choices such as the base action policy and the parameterization of the reward function. Overall, we find that our method robustly produces the policy with the best tradeoff between speed and success.

\subsection{Implementation Details}
\textbf{Imitation Learning Details}. Action Chunking with Transformers (ACT)~\citep{zhao2023learning} models the policy distribution \( p(A_t|s_t) \) as a transformer-based Conditional Variational Autoencoder (CVAE)~\citep{kingma2013auto, sohn2015learning}. The decoder outputs a chunk of actions of size \( k_c \). Chunking is particularly important for high-frequency, fine-grained manipulation tasks, with chunk sizes \( k_c \) being as high as 100~\citep{zhao2023learning}.
3) The input observations $o_t$ includes both proprioception joint positions and camera image embedding through a pretrained image encoder, which enables the algorithm to be effectively used on real robots while not adding too much computing overhead.

\textbf{RL details}
RainbowDQN~\citep{hessel2018rainbow} models the Q function in discrete action space $Q(q_t | s_t, a_t)$, enhancing DQN with several improvements: double Q-learning~\citep{vanhasselt2015deep} reduces overestimation bias, prioritized experience replay~\citep{schaul2015Prioritized} increases sampling efficiency, and the dueling architecture~\citep{wang2016dueling} improves performance in environments where action values are state-independent. Additionally, distributional Q-learning~\citep{bellemare2017distributional} captures reward uncertainty by modeling a categorical distribution over Q-values, making it well-suited for high-dimensional, dynamic tasks in fine-grained manipulation.

% \begin{figure}
%         \centering
%         \includegraphics[width=0.4\textwidth]{figures/ablations/Encoder_Acceleration.png}
%         \includegraphics[width=0.4\textwidth]{figures/ablations/Encoder_Success Rate.png}
%         \caption{Ablation for image encoders}

%         \label{fig:ab_ie}
% \end{figure}

 \begin{figure*}[t!]
        \includegraphics[width=.32\linewidth]{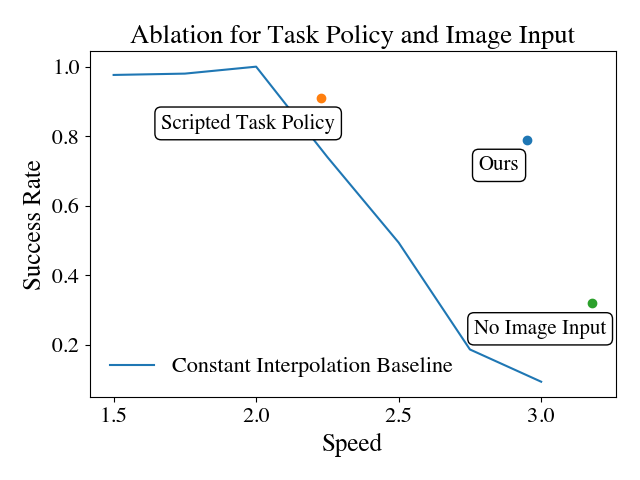}
        \includegraphics[width=.32\linewidth]{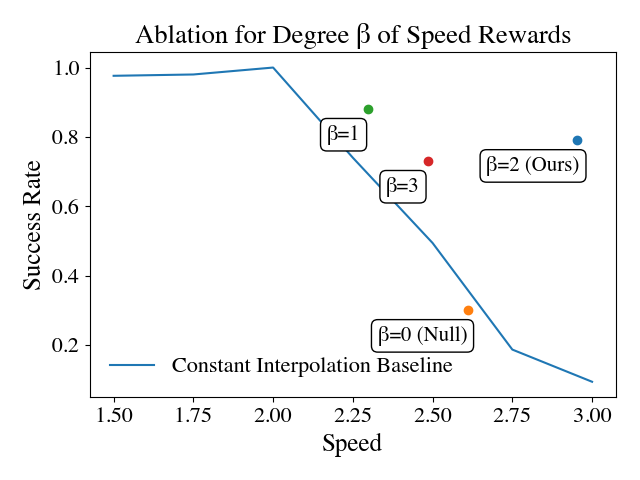}      
        \includegraphics[width=.32\linewidth]{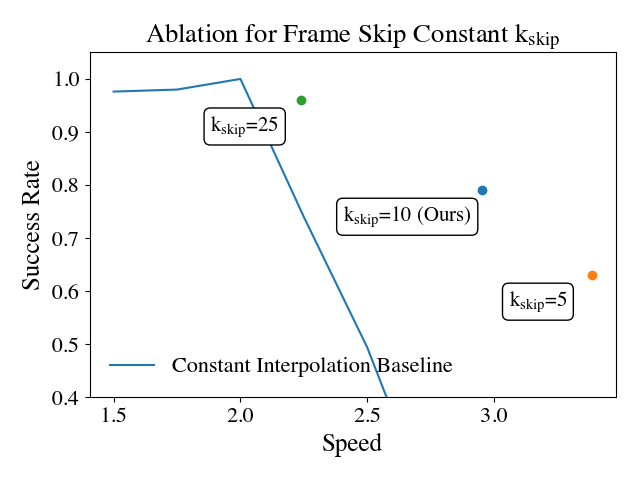}
        \caption{\small Ablation study on task policy, speed rewards, and frame skip values. All configurations are evaluated over 2000 episodes on Simulated Tea Bag Transfer.}
        \label{fig:ab_1}
        \vspace{-0.5cm}
\end{figure*}

\subsection{Robot Setup and Tasks}

We conducted all our experiments based on the ALOHA setup for simulated and real-world tasks. The bimanual hardware setup operates at 50Hz,  consists of a 14-dimensional action space, and the observation states include proprioception (joint position and speed) and 4 RGB cameras. 

We conducted evaluations across six fine-grained manipulation tasks, divided equally between simulated and real-world environments. These tasks encompass long-horizon, contact-rich, and dynamic manipulation scenarios, reflecting the complexity of daily robotic applications.

\textbf{Simulated Cube Transfer:} This task requires the robot to pick up and transfer a cube between two grippers, assessing the system's ability to handle basic object manipulation.

\textbf{Simulated Peg Insertion:} In this task, the robot must pick up and accurately insert a peg into a cube with a hole, demanding precise and bimanual manipulation.

\textbf{Simulated Tea Bag Transfer:} This task involves picking up and transferring a tea bag from a desktop into a cup. The dynamic nature of this task, including the swinging motion of the tea bag during transfer, presents significant challenges for our approach.

\textbf{Tea Bag Disposal:} Extending the simulated tea bag transfer to the real world, this task requires more precise bimanual and dynamic manipulation. The robot must first pick up a mug containing a tea bag and then carefully dispose of the tea bag into a garbage bin.

\textbf{Food Preparation:} This task involves a sequence of precise maneuvers: picking up a plate with one gripper, picking up and throwing two pieces of chicken from the desktop onto the plate, and finally placing the served food at a designated location on the desktop.

\textbf{Pouring Almond:}  This highly precise and dynamic task involves opening the lid of a container and pouring almonds from a can into the container.

\subsection{Key Results}

\textbf{\ours{} significantly accelerates the policy.}
As illustrated in Fig.~\ref{fig:res_sim} and \ref{fig:res_real}, \ours{} achieves over a $2.4\times$ speed-up across all six tasks compared to the original ACT task policy. We observed a trend where higher speed-ups are attained in tasks requiring less precision, such as Simulated Cube Transfer and Simulated Tea Bag Transfer.
This demonstrates that achieving higher speed-ups necessitates a substantial level of precision in task execution.

\textbf{\ours{} effectively retains the success rate at high speed.}
While accelerating the base policy, \ours{} maintains a high success rate even at increased speeds, significantly outperforming the universal interpolation baseline. Notably, \ours{} emerges as an outlying point beyond the Pareto curve defined by the baseline. This trend is particularly pronounced in more dynamic tasks, such as Simulated Tea Bag Transfer, real-world Tea Bag Disposal, and Almond Pouring. This highlights that naive acceleration, as in the baseline, is insufficient for dynamic tasks, requiring a deeper understanding of task-specific dynamics—such as selecting the appropriate speed for transferring a swinging tea bag—something \ours{} effectively accomplishes through reinforcement learning.

\textbf{\ours{} learns the critical parts of the tasks to act accordingly.}
To achieve high acceleration while maintaining success rates, \ours{} intelligently segments the task based on dynamic and contact-rich phases, adjusting its speed accordingly. For instance, as shown in Fig.~\ref{fig:traj_real}, the resulting policy for Almond Pouring demonstrates this capability: \ours{} learns to maintain a high speed during the lid-opening and can-picking phases, deliberately slows down during the transfer and pouring stages to prevent spilling, and then resumes high speed to place the can back.

\section{Ablations}
In this section, we conduct ablation studies on various design choices for \ours{}. All experiments are carried out using Simulated Tea Bag Transfer due to its precise and dynamic nature, as well as its efficiency in simulation.

\textbf{Choice of Task Policy:}
Replacing the learned ACT task policy in \ours{} with an open-loop, scripted policy eliminates the learned generalization capabilities of the task policy. As shown in Fig.~\ref{fig:ab_1}, the scripted policy results in lower acceleration and a slightly higher success rate. This underscores the importance of generalizing across various speeds for the underlying task policy in \ours{}.

\textbf{Image Observations:} 
The speed policy uses both proprioceptive and visual observations for speed predictions. To evaluate whether proprioceptive data alone suffices to represent the environment state, we removed the image input from the policy. Shown in Fig.~\ref{fig:ab_1}, omitting image observations led to less stable training and lower success rates and acceleration. These results highlight the necessity of incorporating visual information for more efficient task execution.

\textbf{Degree of Speed Rewards:}
The speed reward is formulated as $r_\text{speed}(v)= v^\beta$. As shown in Fig.~\ref{fig:ab_1}, varying the degree of the speed reward, $\beta$, has a significant impact on \ours{} performance. When $\beta$ is large, the speed policy prioritizes velocity over task completion, resulting in lower success rates. Conversely, when 
$\beta=0$, there is no direct incentive for speed, and acceleration relies solely on the standard discount factor $\gamma$, leading to unstable training.

\textbf{Frame Skip:} In \ours{}, a frame skip of 10 is used to shorten the horizon of the MDP process. While smaller frame skips provide finer granularity, the extended horizon diminishes the emphasis on terminal success rewards. Conversely, larger frame skips improve short-horizon properties but reduce control precision over speed adjustments. As shown in Fig.~\ref{fig:ab_1}, smaller frame skips with longer horizons lead to suboptimal speed policies, while larger skips sacrifice the granularity needed for achieving higher acceleration.

% \textbf{Image Encoders:} We use a pre-trained ResNet-18 in \ours{} due to its accessibility and strong general-domain performance. Additionally, we tested the image encoder from the ACT task policy, which was fine-tuned for the specific task, and a randomly initialized (non-pretrained) ResNet-18. All three encoders yielded similar final performance, though the random ResNet showed a lower initial success rate before catching up later in training. These results suggest that while general image domain knowledge aids \ours{}, the speed policy does not rely on task-specific knowledge for efficient execution.

\section{CONCLUSION}
We introduced \ours{}, a reinforcement learning framework designed to optimize both execution speed and task success for learned manipulation policies. By training a speed policy to dynamically adjust action velocities based on the current state, \ours{} achieved over 2.4x speed-ups across various dynamic and precise tasks while maintaining adequate success rates compared to baseline methods. Overall, we hope our solution provides valuable insights into the relationship between execution speed and task success in robotic manipulation. Future work should focus on enhancing sample efficiency to make the method more practical for real-world applications.

%%%%%%%%%%%%%%%%%%%%%%%%%%%%%%%%%%%%%%%%%%%%%%%%%%%%%%%%%%%%%%%%%%%%%%%%%%%%%%%%

%%%%%%%%%%%%%%%%%%%%%%%%%%%%%%%%%%%%%%%%%%%%%%%%%%%%%%%%%%%%%%%%%%%%%%%%%%%%%%%%



% \section*{ACKNOWLEDGMENT}

% The preferred spelling of the word ÒacknowledgmentÓ in America is without an ÒeÓ after the ÒgÓ. Avoid the stilted expression, ÒOne of us (R. B. G.) thanks . . .Ó  Instead, try ÒR. B. G. thanksÓ. Put sponsor acknowledgments in the unnumbered footnote on the first page.

%%%%%%%%%%%%%%%%%%%%%%%%%%%%%%%%%%%%%%%%%%%%%%%%%%%%%%%%%%%%%%%%%%%%%%%%%%%%%%%%
\printbibliography

\onecolumn
\appendix[Simulation Rollouts and Additional Ablations]
\captionsetup{skip=3pt}
\setlength{\intextsep}{3pt plus 1pt minus 1pt}
\setlength{\floatsep}{3pt plus 1pt minus 1pt}
\setlength{\textfloatsep}{5pt plus 1pt minus 2pt}

\noindent\begin{minipage}{\linewidth}
\subsection{Simulated Rollout Progressions}
\label{app:simulated_task_progress}

Figure~\ref{fig:appendix_sim_progress} provides a qualitative view of \ours{} on the three simulated tasks. Cube Transfer proceeds from grasping to a bimanual handoff; Peg Insertion requires grasping, alignment, and insertion; and Tea Bag Transfer requires transporting the suspended tea bag into the receptacle. All frames are taken from the original evaluation rollouts.

\par\vspace{3pt}\centering
    \includegraphics[width=.78\linewidth]{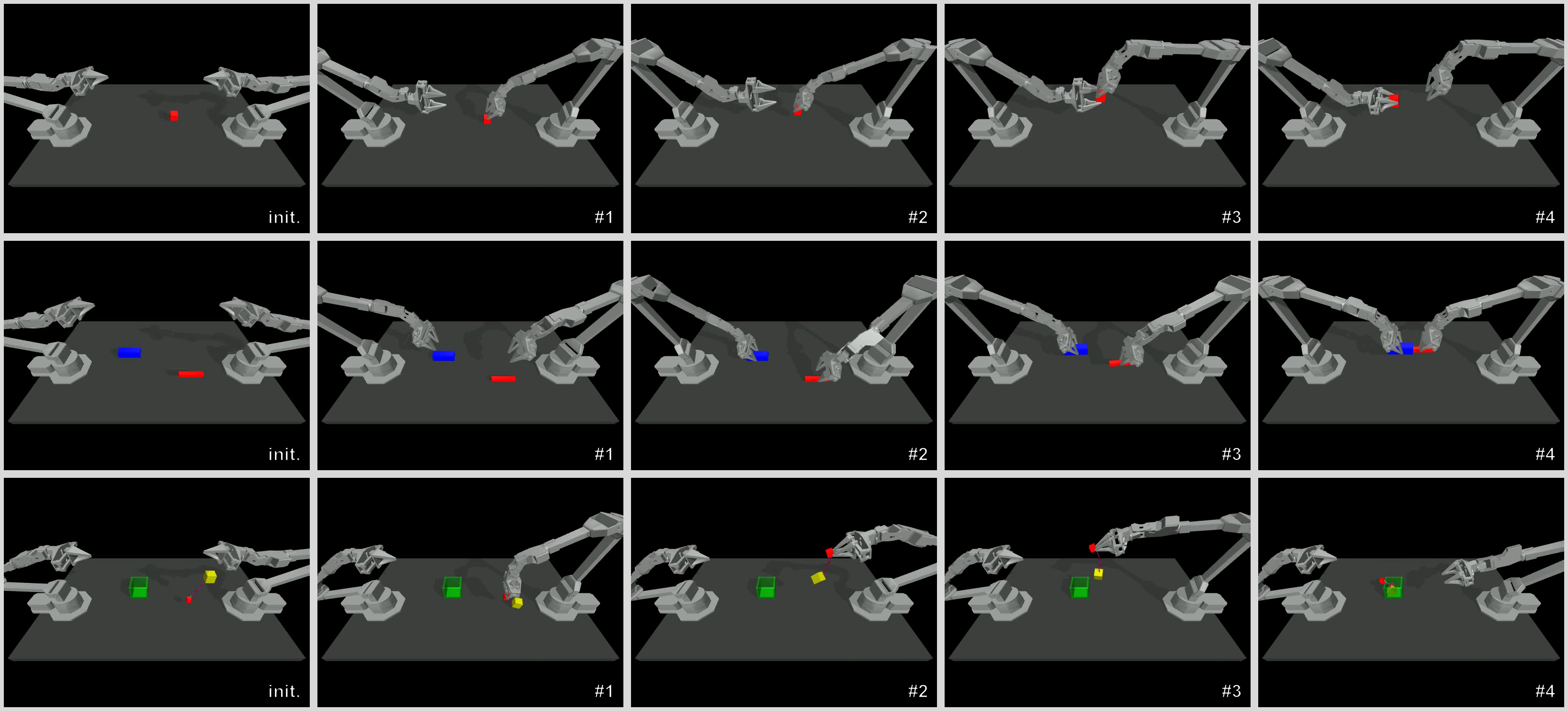}
    \captionof{figure}{Representative \ours{} rollouts in simulation. Rows show Cube Transfer, Peg Insertion, and Tea Bag Transfer; columns show the initial state followed by four successive task stages.}
    \label{fig:appendix_sim_progress}
\end{minipage}

\medskip
\noindent\begin{minipage}{\linewidth}
\subsection{Task Policy}
\label{app:task_policy}

The following plots show the original training curves underlying the endpoint ablations in Fig.~\ref{fig:ab_1}; all configurations are evaluated on Simulated Tea Bag Transfer. In Fig.~\ref{fig:appendix_task_policy}, learned ACT supports substantially higher acceleration, whereas the scripted policy maintains a higher success rate.

\par\vspace{3pt}\centering
    \includegraphics[width=.29\linewidth]{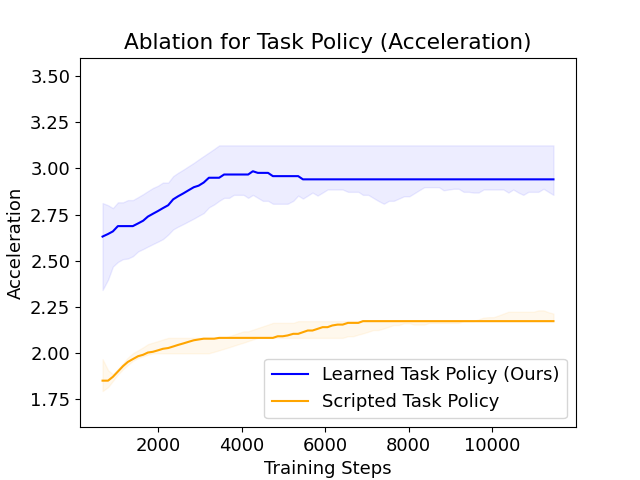}
    \hspace{.025\linewidth}
    \includegraphics[width=.29\linewidth]{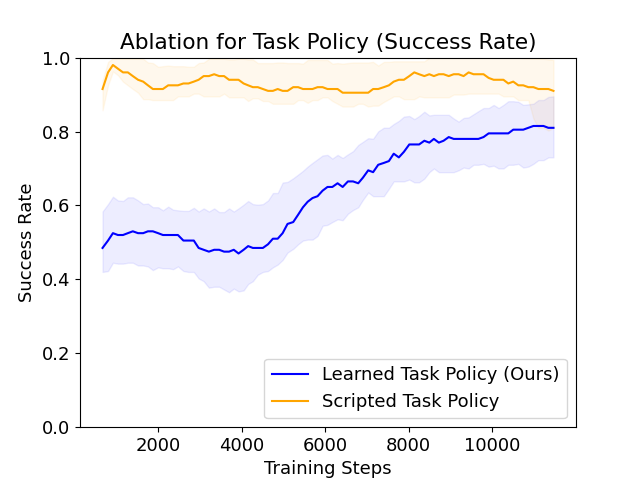}
    \captionof{figure}{Base-policy ablation: acceleration (left) and success rate (right). Learned ACT is compared with an open-loop scripted policy.}
    \label{fig:appendix_task_policy}
\end{minipage}

\medskip
\noindent\begin{minipage}{\linewidth}
\subsection{Image Observations}
\label{app:image_observations}

Removing image observations produces high nominal acceleration but a much lower success rate (Fig.~\ref{fig:appendix_observations}). Visual input is therefore necessary for identifying task phases in which aggressive acceleration would compromise completion.

\par\vspace{3pt}\centering
    \includegraphics[width=.29\linewidth]{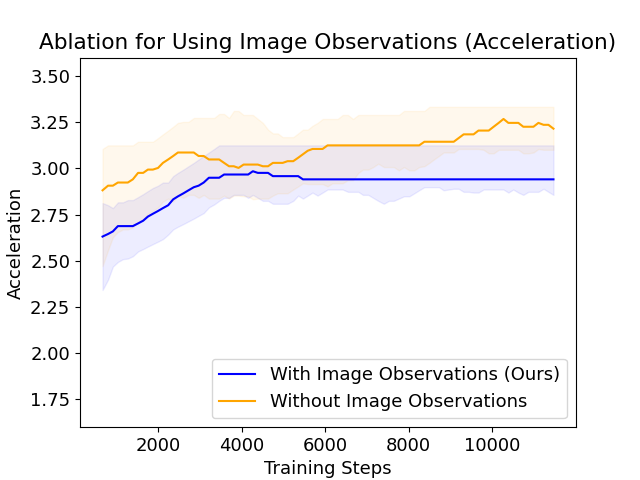}
    \hspace{.025\linewidth}
    \includegraphics[width=.29\linewidth]{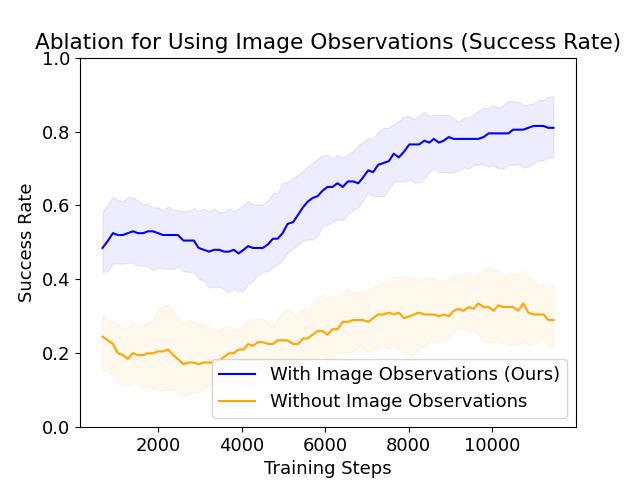}
    \captionof{figure}{Visual-observation ablation: acceleration (left) and success rate (right). Both variants retain proprioception.}
    \label{fig:appendix_observations}
\end{minipage}

\medskip
\noindent\begin{minipage}{\linewidth}
\subsection{Speed-Reward Exponent}
\label{app:speed_reward}

The exponent $\beta$ controls how strongly the reward favors increased speed. Figure~\ref{fig:appendix_reward} shows that $\beta=2$ provides the strongest combined acceleration and final success; $\beta=0$ is less stable, while $\beta=1$ and $\beta=3$ favor lower-acceleration operating points.

\par\vspace{3pt}\centering
    \includegraphics[width=.29\linewidth]{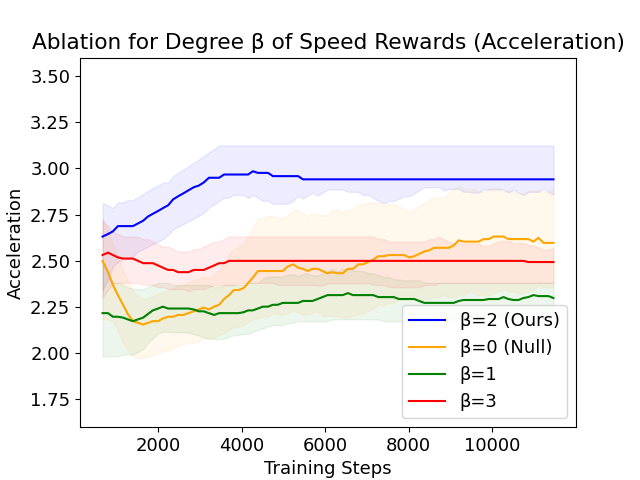}
    \hspace{.025\linewidth}
    \includegraphics[width=.29\linewidth]{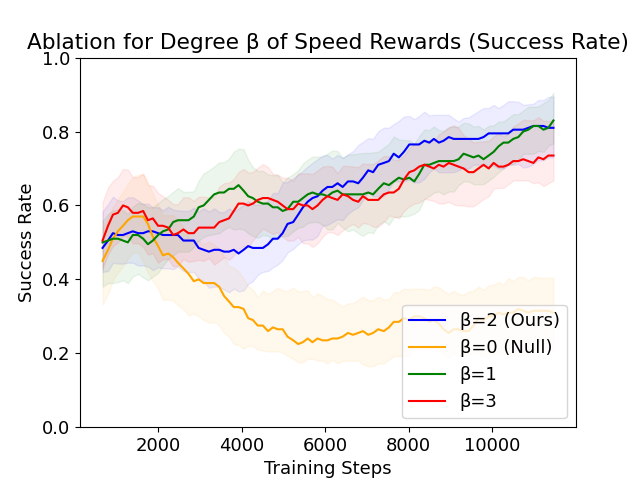}
    \captionof{figure}{Speed-reward ablation: acceleration (left) and success rate (right). The main experiments use $\beta=2$.}
    \label{fig:appendix_reward}
\end{minipage}

\medskip
\noindent\begin{minipage}{\linewidth}
\subsection{Frame Skip}
\label{app:frame_skip}

Frame skip $f_s$ sets the interval between speed-policy decisions. In Fig.~\ref{fig:appendix_frameskip}, $f_s=5$ favors acceleration, $f_s=25$ favors success, and the selected value $f_s=10$ balances the two objectives.

\par\vspace{3pt}\centering
    \includegraphics[width=.29\linewidth]{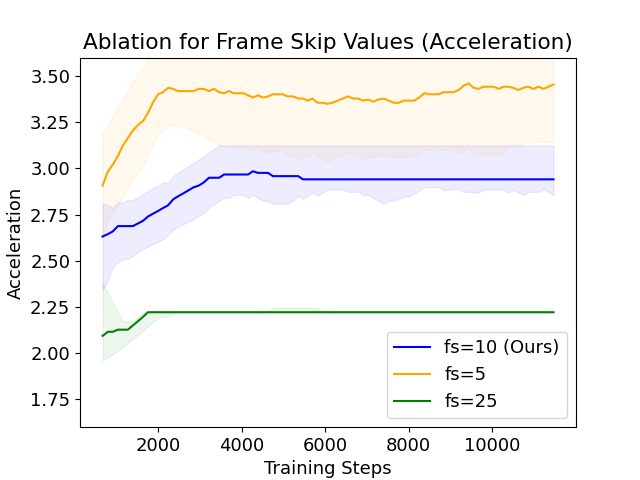}
    \hspace{.025\linewidth}
    \includegraphics[width=.29\linewidth]{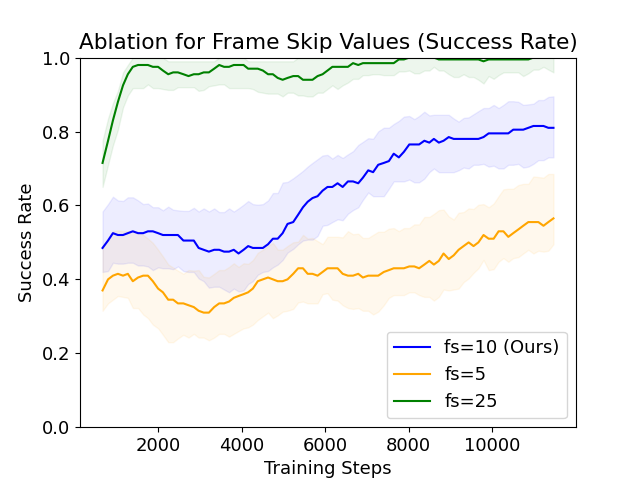}
    \captionof{figure}{Frame-skip ablation: acceleration (left) and success rate (right). The main experiments use $f_s=10$.}
    \label{fig:appendix_frameskip}
\end{minipage}

\medskip
\noindent\begin{minipage}{\linewidth}
\subsection{Image Encoder}
\label{app:image_encoder}

All three encoders reach similar final acceleration (Fig.~\ref{fig:appendix_encoder}). The randomly initialized encoder learns success more slowly, while the pretrained ResNet-18 attains the highest final success, showing that the speed policy need not reuse the task-policy encoder.

\par\vspace{3pt}\centering
    \includegraphics[width=.29\linewidth]{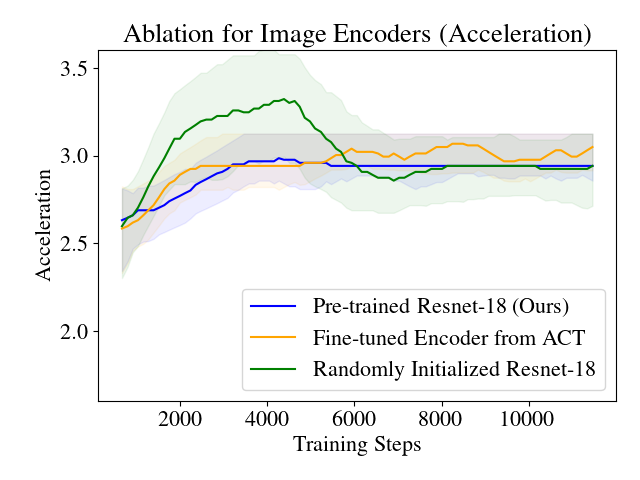}
    \hspace{.025\linewidth}
    \includegraphics[width=.29\linewidth]{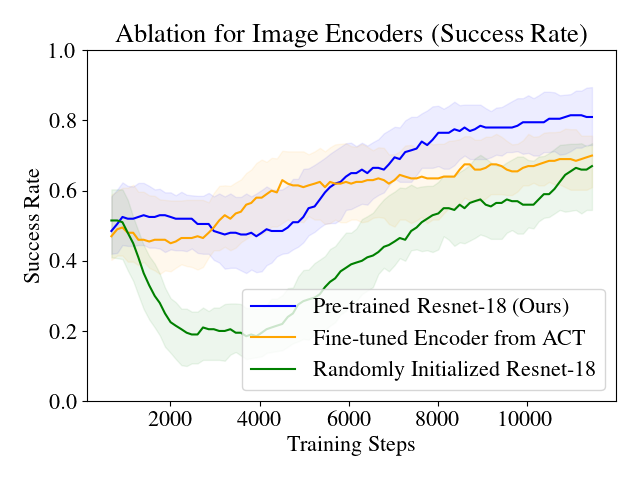}
    \captionof{figure}{Image-encoder ablation: acceleration (left) and success rate (right).}
    \label{fig:appendix_encoder}
\end{minipage}

\end{document}